\documentclass[letterpaper, 10 pt, conference]{ieeeconf}

\IEEEoverridecommandlockouts                         
\usepackage{bm}
\usepackage{amssymb}
\usepackage{threeparttable}
\usepackage{wrapfig}
\usepackage{amsmath}
\usepackage{wrapfig}
\usepackage{lipsum}
\usepackage{siunitx}
\usepackage{booktabs}
\usepackage{multirow}
\usepackage{glossaries}
\usepackage{hyperref}
\usepackage{graphicx}
\usepackage{url}
\usepackage{cleveref}
\usepackage{cite}
\usepackage{times}
\usepackage[table]{xcolor}

\title{\LARGE \bf Ubiquitous Robot Control Through Multimodal Motion Capture Using Smartwatch and Smartphone Data}

\author{
Fabian C Weigend$^{1}$, 
Neelesh Kumar$^{2}$,
Oya Aran$^{2}$,
and
Heni Ben Amor$^{1}$
\thanks{$^{1}$Weigend, F \& Ben Amor, H are affiliated with SCAI, Arizona State University \texttt{$\lbrace$fweigend, hbenamor$\rbrace$@asu.edu}.
$^{2}$Kumar, N \& Aran, O. are affiliated with Corporate Functions-R\&D, Procter and Gamble
\texttt{$\lbrace$kumar.n.40,aran.o$\rbrace$@pg.com}
}}

\begin{document}

\maketitle
\thispagestyle{empty}
\pagestyle{empty}

\begin{abstract}
We present an open-source library for seamless robot control through motion capture using smartphones and smartwatches. Our library features three modes: Watch Only Mode, enabling control with a single smartwatch; Upper Arm Mode, offering heightened accuracy by incorporating the smartphone attached to the upper arm; and Pocket Mode, determining body orientation via the smartphone placed in any pocket.
These modes are applied in two real-robot tasks, showcasing placement accuracy within 2\,cm compared to a gold-standard motion capture system. WearMoCap stands as a suitable alternative to conventional motion capture systems, particularly in environments where ubiquity is essential. The library is available at: \url{www.github.com/wearable-motion-capture}.
\end{abstract}

\section{INTRODUCTION}

As consumer wearables become increasingly ubiquitous, IMU-based motion capture utilizing smartwatch and smartphone data emerges as a widely accessible solution~\cite{imuPoser,miaomiao2023}. While motion capture from consumer wearables may offer lower accuracy compared to optical and specialized IMU-based systems, their key advantage lies in their ubiquity and familiarity. 
Most users carry these devices with them consistently, enabling pose tracking anytime and anywhere~\cite{imuPoser}.

Recently, we showcased the potential of utilizing smartwatch data for robot control~\cite{weigend2023anytime,weigend2024iroco}. In this work, we compile these approaches into a library named Wearable Motion Capture (WearMoCap). It offers three operational modes tailored to different requirements of precision and portability. We evaluate WearMoCap in two real-robot tasks.

\section{WearMoCap MODES}

\label{subsec:mocap_modes}
We begin by outlining the pose estimation methodology for the three motion capture modes as depicted in \Cref{fig:teaser}. 

\subsubsection{Watch Only} 
Utilizing the optimal neural network architecture proposed by \cite{weigend2023anytime}, an LSTM is employed to estimate the orientations of the lower arm, ${\bf q}_\mathrm{la}$, and the upper arm, ${\bf q}_\mathrm{ua}$, derived from a sequence of watch sensor data, including calibrated IMU, orientation, and pressure readings. Subsequently, positional values are computed through forward kinematics. The Watch Only mode requires the user to maintain a constant forward-facing direction post-calibration.

\subsubsection{Upper Arm} 
\begin{figure}
    \centering
    \includegraphics[width=1\linewidth]{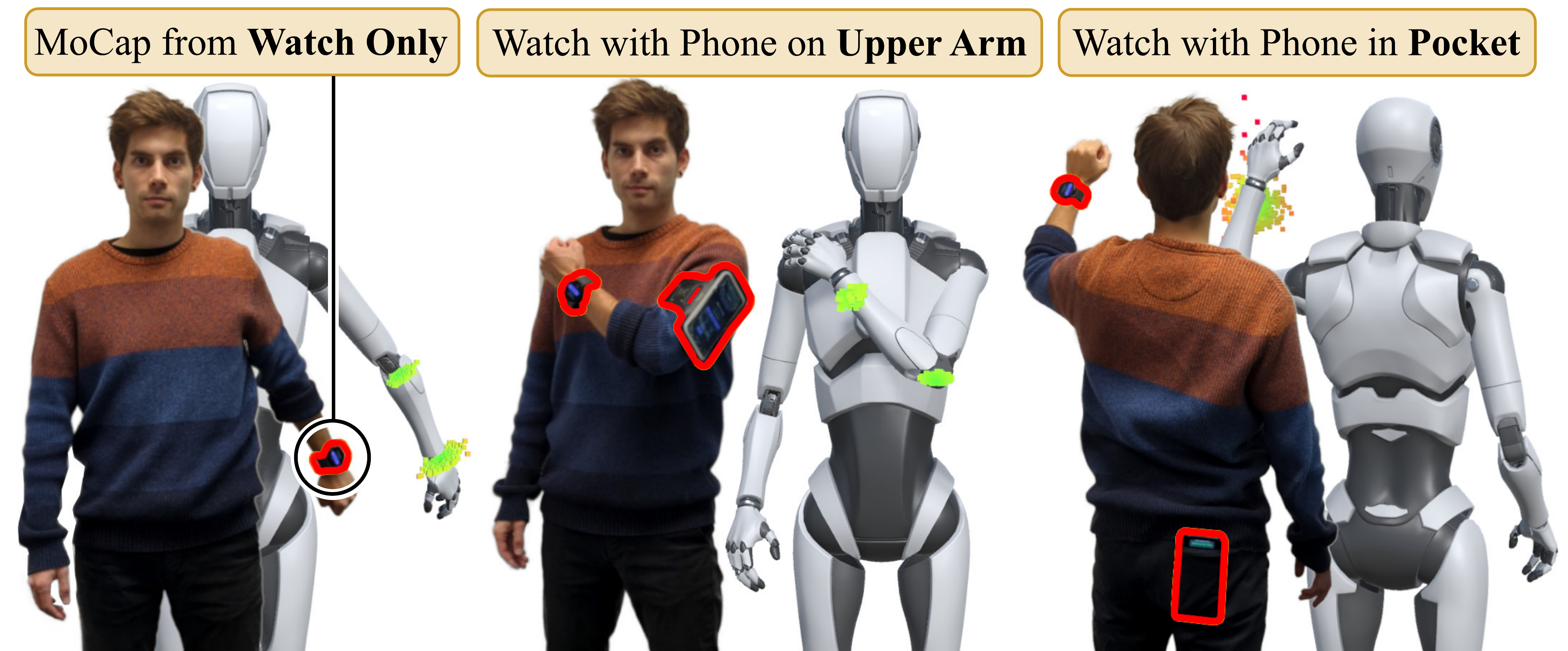}
    \caption{Our three WearMoCap modes track the arm pose with varying accuracy and flexibility. We evaluated all modes in real-robot tasks.}
    \label{fig:teaser}
    \vspace{-0.2in}
\end{figure}

In this mode, the user wears a watch and straps a phone to their upper arm (see \Cref{fig:teaser}). Utilizing an LSTM with three layers of 128 neurons, WearMoCap estimates ${\bf q}_\mathrm{la}$ and ${\bf q}_\mathrm{ua}$ from the combined sensor data from both devices. Joint positions are then estimated through forward kinematics. In contrast to Watch Only mode, users are free to turn and change their body orientation.

\subsubsection{Pocket} This mode streams the sensor data from the watch and a phone in the user's pocket. WearMoCap employs a Differentiable Ensemble Kalman Filter \cite{liu2023enhancing} to update an ensemble of states from previous estimates and the sensor data from both devices. For the detailed methodology, we refer to \cite{weigend2024iroco}. Each ensemble member characterizes the orientation of the lower arm, ${\bf q}_\mathrm{la}$, the upper arm, ${\bf q}_\mathrm{ua}$, and additionally captures the rotation around the up-axis of the hip, ${\bf q}_\mathrm{hi}$. Joint positions are determined through forward kinematics.

\section{REAL-ROBOT TASKS}
\label{sec:real-robot-tasks}
We assessed the practicality of WearMoCap in two human-robot tasks approved by the ASU IRB as STUDY00018521.

\subsection{Handover}

\begin{figure}
    \centering
    \includegraphics[width=0.92\linewidth]{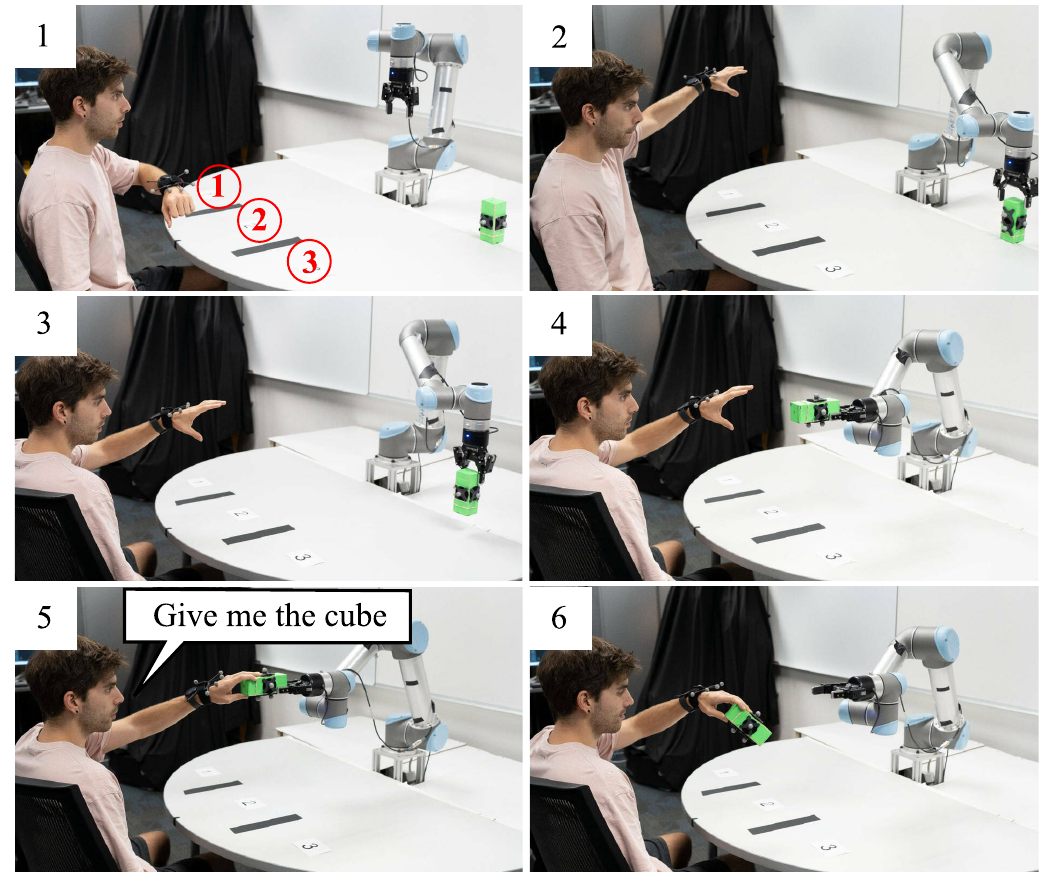}
    \caption{\textbf{Handover Task:} Human subjects used WearMoCap to perform handovers with a UR5. A voice command completed the handover and the robot let go of the cube.}
    \label{fig:handover}
\end{figure}

In the handover task, a Universal Robot 5 (UR5) picked up an object from a table and handed it over to a human subject. Successful completion of this task relied on WearMoCap accurately tracking the position of the human hand. Human subjects were seated on a rotating chair facing the UR5. As illustrated in Step~1 of \Cref{fig:handover}, the tabletop area between the robot and the subject was divided into three distinct zones. Within these zones, subjects performed two handovers: one with the hand positioned at a lower and another with the hand at a higher elevation. Subjects repeated these handovers in all other zones in a random order. While the subject's orientation remained fixed in the Watch Only mode, they had the flexibility to adjust their orientation by rotating the chair in the other two modes.

To mitigate potential learning effects, we randomized the order of tracking modes. We recorded the handover distance as the disparity between the hand position and the cube when the participant completed the handover with a voice command (Step~5). Additionally, we recorded handover time, representing the duration taken from Step~2 to Step~5.

\begin{figure}
    \centering
    \includegraphics[width=0.92\linewidth]{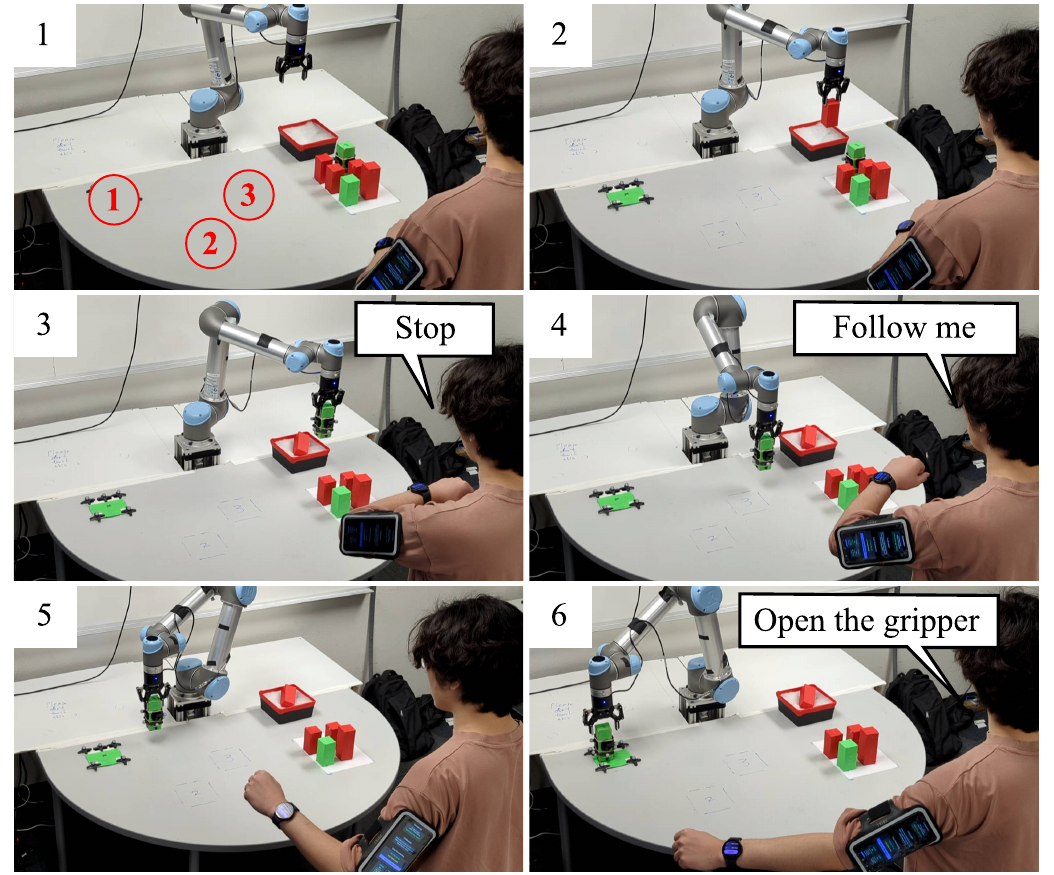}
    \caption{\textbf{Intervention Task:} The UR5 sorted the green and red cubes into the bin. Human subjects use WearMoCap to interrupt the robot and to place the green cube at a target location.}
    \label{fig:intervention}
    \vspace{-0.1in} 
\end{figure}

\subsection{Intervention}

For this task, the UR5 robot autonomously picked up colored cubes (green or red) and dropped them at target locations of the corresponding color. As depicted in \Cref{fig:intervention} Step~1, the human subject stood facing the robot, and there were three possible target locations for the green cube. However, the robot was not trained to correctly place green cubes. Whenever the robot selected a green cube, the subject intervened with a voice command and directed the robot with WearMoCap to place the cube correctly.

Five subjects performed this task with each WearMoCap mode randomly assigned and for each of the three green target locations. We used OptiTrack to record the placement distance as the distance between the placed green cube and the center of its target location. Additionally, we recorded the task completion time as the duration from issuing the "Follow me" command in Step~4 until Step~5.

\section{Results And Conclusions}

As our baseline, subjects completed all tasks using OptiTrack. \Cref{Tab:robot_applications} compares baseline objective task metrics to the performance achieved with WearMoCap modes. Notably, the Watch Only mode exhibits higher error rates, as indicated by its greater intervention placement distance (+5.2\,cm) and handover distance (+4.5\,cm). Conversely, the Upper Arm mode demonstrates superior accuracy, with deviations below +2\,cm in both tasks. Also, the Pocket mode surpasses the Watch Only mode in distance metrics, attributable to its additional degree of freedom for fine-tuning positioning. However, this extra flexibility resulted in longer task completion times, as subjects had to navigate adjustments in arm motion alongside changes in body orientation.

\begin{table}[ht]
\begin{center}
\scriptsize
\caption{Summarized Differences to OptiTrack Baseline}
\label{Tab:robot_applications}
\begin{tabular}{c r S[table-format=3.1] @{${}\pm{}$} S[table-format=1.1] S[table-format=3.1] @{${}\pm{}$} S[table-format=2.1] c}
\toprule 

Task & Method & \multicolumn{2}{c}{Dist. (cm)} & \multicolumn{2}{c}{Time (s)} &  Trials\\

\midrule 
\multirow{4}{1em}{\rotatebox[origin=c]{90}{Handover}} 
 & \textbf{OptiTrack} & 6.8 & 1.6 & 9.2 & 3.2 & 24\\
 & Watch Only   & {+}4.5 & 9.7 & {+}3.3 & 8.1 & 24\\
 & Pocket & {+}2.2 & 3.6 & {+}3.4 & 6.3 & 24\\
 & Upper Arm   & {+}1.9 & 3.7 & {+}0.5 & 5.2 & 24\\

\midrule 

\multirow{4}{1em}{\rotatebox[origin=c]{90}{Intervent.}} 
 & \textbf{OptiTrack} & 2.4 & 1.5 & 17.5 & 4.5 & 15 \\
 & Watch Only   & {+}5.2 & 6.0 & {+}10.5 & 7.8 & 15 \\
 & Pocket & {+}2.9 & 4.3 & {+}11.4 & 10.7 & 15 \\
 & Upper Arm   & {+}1.7 & 5.2 & {+}4.0 & 5.6  & 15 \\

\bottomrule
\end{tabular}
\end{center}
\vspace{-0.1in}
\end{table}

Overall, WearMoCap proved suitable for the evaluated real-robot tasks. However, variations in task performance among the three modes revealed application trade-offs.

\paragraph{Watch Only} Solely utilizing a smartwatch offers convenience in terms of availability and setup. However, our findings exhibit considerable increases in placement deviations and completion times. The suitability of this mode depends on the task at hand; while the human can mitigate final centimeter deviations during handovers, the prediction inaccuracies are impractical for high-fidelity teleoperation control, such as pick-and-place tasks.

\paragraph{Upper Arm} Wearing an upper-arm fitness strap is an additional step compared to the other modes. Yet, our results confirm that an additional IMU on the upper arm enables accurate pose tracking, as it has been previously established by \cite{yang2016neural,joukov2017human}. With the widespread availability of smart devices, this mode emerges as a ubiquitous and promising alternative to OptiTrack for robot teleoperation.

\paragraph{Pocket} The Pocket mode simply requires users to stow the phone in a pocket and allows to move around freely. This stands in contrast to the Watch Only mode, where users must maintain a fixed forward-facing direction. Our results underscore that the additional tracking of body orientation facilitated more precise control than in Watch Only mode. Still, results are less precise than with the Upper Arm mode. The Pocket mode, therefore, balances the precision and convenience of the other two modes.

In conclusion, WearMoCap stands as a promising alternative to state-of-the-art motion capture systems. When weighing its trade-offs for suitable tasks, it effectively transforms smartwatches and smartphones into versatile interfaces, facilitating intuitive robot control anytime and anywhere.

\bibliographystyle{IEEEtran}
\scriptsize{
\bibliography{references}
}

\end{document}